\documentclass[runningheads]{llncs}
\usepackage[T1]{fontenc}
\usepackage{graphicx}

\usepackage{color}

\usepackage{xcolor} 
\usepackage{subcaption}
\usepackage[backref=page]{hyperref}
\usepackage{amsmath,amssymb}
\hypersetup{
    colorlinks=true,
    linkcolor=blue,
    filecolor=magenta,      
    urlcolor=cyan,
    pdftitle={Paper Title},
    pdfpagemode=FullScreen,
}
\usepackage{multirow}
\usepackage{float}
\usepackage{placeins}

\begin{document}

    \title{Neural Descriptors: Self-Supervised Learning of Robust Local Surface Descriptors Using Polynomial Patches}
    \titlerunning{Neural Descriptors}
    %
    \author{
    Gal Yona \orcidID{0009-0002-8519-8230} \and
    Roy Velich\orcidID{0009-0005-3111-0055} \and
    Ehud Rivlin\orcidID{0009-0008-6432-9127} \and
    Ron Kimmel\orcidID{0000-0002-3180-7961}
    }
    \authorrunning{G. Yona et al.}
    
    %
    \institute{Technion - Israel Institute of Technology}
    %
    \maketitle              

    \begin{abstract}
        Classical shape descriptors such as Heat Kernel Signature (HKS), Wave Kernel Signature (WKS), and Signature of Histograms of OrienTations (SHOT), while widely used in shape analysis, exhibit sensitivity to mesh connectivity, sampling patterns, and topological noise. While differential geometry offers a promising alternative through its theory of differential invariants, which are theoretically guaranteed to be robust shape descriptors, the computation of these invariants on discrete meshes often leads to unstable numerical approximations, limiting their practical utility.
        We present a self-supervised learning approach for extracting geometric features from 3D surfaces. Our method combines synthetic data generation with a neural architecture designed to learn sampling-invariant features. By integrating our features into existing shape correspondence frameworks, we demonstrate improved performance on standard benchmarks including FAUST, SCAPE, TOPKIDS, and SHREC'16, showing particular robustness to topological noise and partial shapes.
    
        \keywords{3D Shape Analysis \and Geometric Invariants \and Surface Representation.}
    \end{abstract}

    \section{Introduction}
        Finding correspondences between deformable 3D shapes is a fundamental problem in computer vision and graphics with numerous applications, from texture and deformation transfer \cite{DinhTextureTrans2005,BaranDeformationTrans2009} to statistical shape analysis and 3D object recognition \cite{PishchulinShapeScape2017,BelongieShapeMatchingObjReco2002}. Despite decades of research, establishing reliable dense correspondences between shapes remains challenging, particularly when dealing with non-rigid deformations, partial observations, and varying mesh connectivity \cite{bracha2024unsupervisedpartialshapecorrespondence,bracha2024wormholelosspartialshape}.
        
        A significant breakthrough came with the functional maps framework \cite{FunctionalMaps2012}, reformulating point-to-point matching as a linear map between functional spaces. This framework has evolved through deep learning approaches, from supervised methods like Deep FM-Net \cite{DeepFunctionalMaps2017} to unsupervised techniques using metric regularization \cite{UnsupervisedLearningDenseShapeCorr2019} and spectral regularization \cite{Roufosse2019,Cao2022}. However, these methods still rely heavily on the quality of input geometric features.
        
        Traditional approaches for characterizing local geometry include spectral descriptors like Heat Kernel Signature (HKS) \cite{HKS2009} and Wave Kernel Signature (WKS) \cite{WKS2011}, which use heat diffusion and quantum mechanics respectively, and histogram-based descriptors like Signature of Histograms of OrienTations (SHOT) \cite{SHOT2014}. While mathematically elegant, these descriptors are sensitive to surface discretization, sampling patterns, and topological noise.
        
        The mathematical theory of differential invariants \cite{cartan1935,olver_1995} offers an alternative framework through quantities unchanged under transformations. For surfaces in 3D, these include principal curvatures and their derivatives along principal directions \cite{Olver2018NormalForms,OLVER2009}. However, computing these on discrete geometry requires estimating high-order derivatives, leading to numerical instabilities. Previous efforts \cite{VelichKimmel2022,VelichKimmel2023} showed that neural networks can overcome such instabilities for planar curves through self-supervised learning, motivating our exploration in 3D.
        
        We propose a self-supervised learning framework for extracting robust geometric features from 3D surfaces. Our key innovation lies in combining synthetic training data with a neural architecture that learns sampling-invariant features while maintaining discriminative power.
        When integrated into shape correspondence frameworks, our features significantly outperform traditional descriptors across multiple benchmarks, particularly for partial shapes and topological noise.
        
        Our main contributions include:
        \begin{enumerate}
            \item A synthetic data generation pipeline creating local surface patches with controlled geometric properties and varying sampling patterns.
            \item A self-supervised neural network architecture learning sampling-invariant geometric features.
            \item State-of-the-art performance on multiple shape matching benchmarks including FAUST, SCAPE, TOPKIDS, and SHREC'16, demonstrating particular robustness to topological noise and extreme isometric deformations.
        \end{enumerate}

         Our code and model weights are available at \href{https://github.com/GalYona21/Neural-Descriptors}{github.com/GalYona21/Neural-Descriptors}.

\section{Related Work}
    \subsubsection{Classical Geometric Descriptors.}
        Numerous approaches have been proposed for characterizing local geometry of 3D shapes. The Heat Kernel Signature (HKS) \cite{HKS2009} uses heat diffusion to capture intrinsic geometry, demonstrating strong performance in capturing both local and global shape features, while its scale-invariant extension (SI-HKS) \cite{SIHKS2010} achieves scale invariance through logarithmically sampled scale-space and Fourier analysis \cite{5995486}, improving stability under deformations. The Wave Kernel Signature (WKS) \cite{WKS2011} provides a complementary perspective by analyzing quantum mechanical particles on the surface, offering better frequency localization than HKS. The Signature of Histograms of OrienTations (SHOT) \cite{SHOT2014} takes a different approach, building a local reference frame and accumulating spatial histograms of normal directions. While these classical descriptors offer elegant mathematical formulations, they exhibit significant sensitivity to discretization and topological noise. Moreover, spectral descriptors face fundamental limitations when attempting to increase their discriminative power: the number of available Laplacian eigenfunctions is bounded by the mesh vertex count, creating an upper limit on feature dimensionality, and higher-frequency eigenfunctions exhibit increased sensitivity to noise and sampling variations.

    \subsubsection{Differential Invariants and Signatures.}
        Differential invariants characterize shapes through quantities unchanged under transformations. For surfaces in 3D, Euclidean invariants involve Gaussian and mean curvatures and their derivatives with respect to the principal directions \cite{Olver2018NormalForms}, while equi-affine signatures use the Pick invariant \cite{OLVER2009}. While Cartan \cite{cartan1935} established the theory and Olver \cite{olver_1995} proved all invariants derive from a single fundamental invariant, computing these quantities faces numerical challenges. Recent work \cite{VelichKimmel2022,VelichKimmel2023} showed neural networks can effectively approximate differential invariants for planar curves, motivating extension to 3D analysis.
        Our method draws inspiration from these axiomatic differential signatures, applying self-supervised learning to capture analogous invariance properties. Rather than directly computing unstable differential quantities, we learn features of local surface patches that remain consistent across different sampling patterns of the same underlying geometry. The design of our approach is guided by the theoretical guarantees of differential invariants—transformation invariance and geometric distinctiveness—which we aim to implicitly encode in the learned features, while avoiding the numerical instabilities associated with their direct computation.

    \subsubsection{Spectral Methods and Learning on 3D Shapes.}
        The functional maps framework \cite{FunctionalMaps2012} reformulated shape correspondence as linear maps between functional spaces. Deep learning approaches like Deep FM-Net \cite{DeepFunctionalMaps2017} and GeomFmaps \cite{DeepGeometricFunctionalMaps2020} learned to transform descriptors, while DiffusionNet \cite{DiffusionNet2022} introduced \\
        discretization-agnostic architectures. However, these methods still struggle with holes and partial shapes due to reliance on Laplace-Beltrami eigenfunctions \cite{bracha2024unsupervisedpartialshapecorrespondence}.
        Our work addresses this by providing robust descriptors that enhance feature extractors like DiffusionNet, particularly improving performance where traditional spectral approaches falter.

    \section{Method}
        Our goal is to train a feature extractor that is both sampling-invariant and discriminative: it should output similar features for different samplings of the same 3D surface while distinguishing between geometrically different surfaces.
        
        Formally, our architecture takes a point cloud $\left\{p_i \in \mathbb{R}^3 \right\}_{i=1}^{N}$ as input and outputs a feature vector $v \in \mathbb{R}^D$ capturing the local surface geometry. Given two different samplings $\left\{p_i \right\}_{i=1}^{N_1}$ and $\left\{q_i \right\}_{i=1}^{N_2}$ of the same surface, their corresponding feature vectors $v_1$ and $v_2$ should be similar (as measured by cosine similarity). This naturally leads us to adopt a contrastive learning approach, where we optimize an objective that encourages feature similarity for different samplings of the same surface.

        
        Therefore, our approach consists of three main components: synthetic data generation of 3D surface patches, feature extraction using DeltaConv \cite{DeltaConv2022}, and self-supervised training using SimSiam \cite{SimSiam2021}.
        In the rest of the paper, we will refer to our learned features as \textit{Neural Descriptors}.
    
        \subsection{Synthetic Data Generation}
            We generate synthetic training patches by sampling random polynomial height fields over a 2D parametric domain $\Omega \subset \mathbb{R}^2$ centered at the origin. For each patch, we select a random polynomial of order $d \in \mathbb{N}$ and coefficients $\{a_{ij}\} \subset \mathbb{R}$ to define:
            
            \begin{equation*}
                f(x,y) = \sum_{i+j\leq d} a_{ij}x^iy^j, \quad (x,y) \in \Omega.
            \end{equation*}
            
            Rather than using a regular grid, we sample a random number of points $N$ uniformly from the continuous domain $\Omega$, yielding point coordinates $\{(x_i, y_i)\}_{i=1}^N$. Each point's $z$-coordinate is computed by evaluating $f(x_i,y_i)$, resulting in the point cloud $\{(x_i, y_i, f(x_i,y_i))\}_{i=1}^N$. This continuous sampling approach ensures good surface coverage while avoiding grid artifacts that could bias learning.
            
            For training, we generate pairs of point clouds from each polynomial surface by varying both the number of points and the sampling pattern. These pairs, representing different discretizations of the same underlying surface, serve as training examples in our contrastive learning framework.
        
            
            

            
            \subsection{Network Architecture}
            We choose DeltaConv \cite{DeltaConv2022} as our backbone architecture, which has demonstrated impressive performance in learning implicit 3D features across various point cloud tasks. While DeltaConv typically learns these features implicitly through task-specific training, we adapt its architecture and explicitly train it to capture geometric features of local 3D patches.
            Unlike feature extractors that are based on spectral analysis requiring pre-computation of the Laplace-Beltrami eigendecomposition \cite{DiffusionNet2022}, DeltaConv operates directly in the spatial domain using geometric operators from vector calculus (gradient, divergence, curl, and Hodge Laplacian). This spatial approach makes it naturally robust to varying sampling patterns and more efficient to compute. Empirically, we found that DeltaConv achieves better performance compared to other state-of-the-art spatial feature extractors, such as point-transformer architectures \cite{PointTransformer2021}.
            
\begin{figure}[t]
            \centering
            \includegraphics[width=\textwidth]{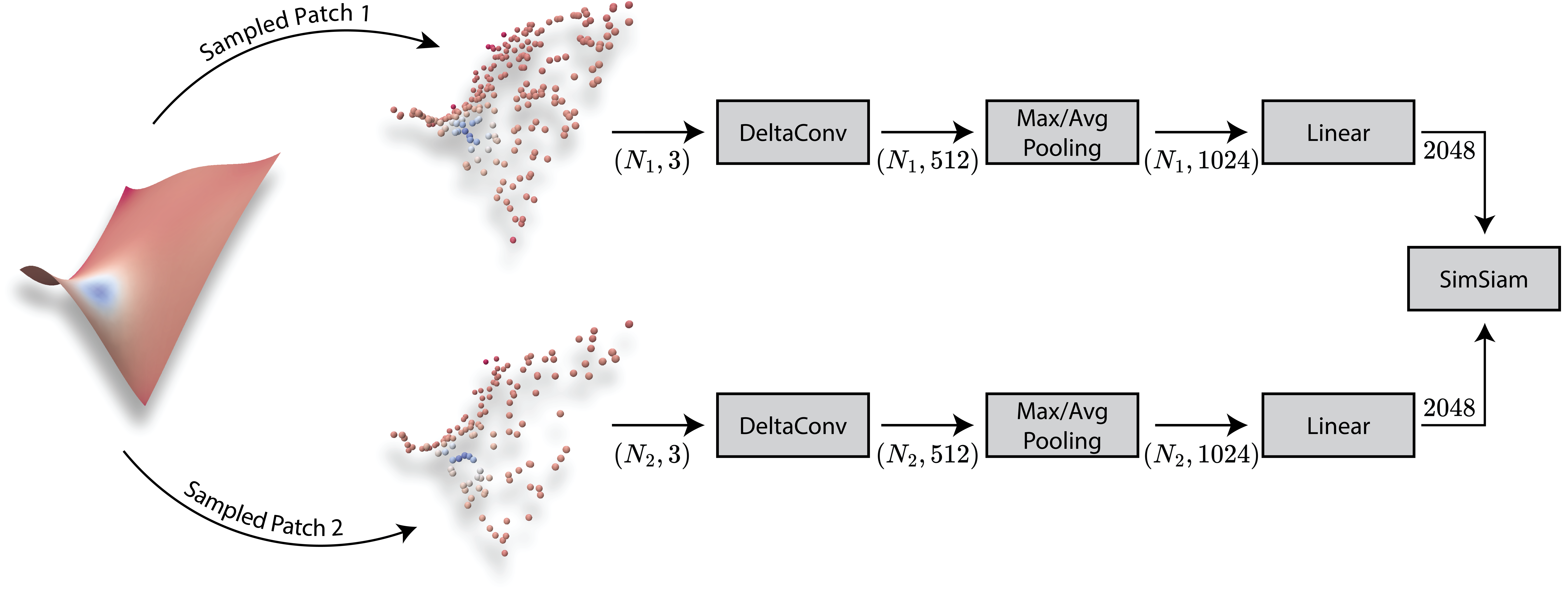}
            \caption{Architecture overview.}
            \label{fig:arch}
        \end{figure}

            Our implementation processes batches of 3D point clouds with varying numbers of points, utilizing Torch Geometric \cite{torch_geometric}\footnote{\href{https://pytorch-geometric.readthedocs.io/en/latest/}{Torch Geometric Website}} to handle these non-homogeneous batches efficiently. The network consists of 4 DeltaConv blocks with increasing feature dimensions (64, 64, 128, 256). We concatenate the outputs of all 4 blocks to obtain a feature vector of 512 dimensions per point in the input point cloud. To obtain a single feature vector that represents the entire patch, we apply both mean and max global pooling operations over all points in the patch. These pooled features are then passed through a linear layer to produce the final 2048 dimensional feature vector for the patch (See Figure \ref{fig:arch}). This final linear layer provides flexibility to adjust the feature dimensionality to suit the inputs required by downstream task models and applications.

        \subsection{Self-Supervised Training with SimSiam}
            We train our feature extractor using SimSiam \cite{SimSiam2021}, a simple and effective self-supervised learning framework. Unlike other contrastive learning approaches, SimSiam does not require negative training samples or momentum encoders, making it computationally efficient and easy to implement. The key idea behind SimSiam is to maximize the similarity between different views of the same data while preventing collapsed solutions through a stop-gradient operation.
            
            During training, we construct batches of 256 pairs of corresponding sampled patches, where each pair consists of two different samplings of the same underlying surface. Each patch is first aligned to a canonical pose through SVD decomposition of the covariance matrix of its points to ensure consistent orientation before being fed into the DeltaConv-based feature extraction pipeline.
            
            The network weights are updated via backpropagation, guided by SimSiam's contrastive learning objective. This objective naturally encourages the network to learn sampling-invariant features by requiring similar feature vectors for different samplings of the same surface, while the framework's stop-gradient mechanism prevents trivial solutions.


      
            The validation process follows a similar structure, using batches of 256 corresponding patch pairs. Each patch is processed through the network to obtain its feature vector. We then compute the cosine similarity between each source patch (first patch in each pair) and all target patches (all the second patches among all pairs) in the batch. For each source patch, we identify which target patch yields the highest cosine similarity between the feature vectors, and record whether this match corresponds to the correct pair. Our training converged when the matching accuracy saturated at approximately 87\% for batches of 256 pairs, meaning that 87\% of source patches were correctly matched with their corresponding target patches from the same underlying surface, despite different sampling patterns. This indicates that our learned features successfully capture consistent geometric information across varying discretizations.

    \section{Experiments}
        \subsection{Qualitative Analysis of Feature Space}
            We conduct a series of experiments to examine the interpretability and geometric consistency of our learned features compared to traditional descriptors (SHOT, HKS, and WKS). Using Principal Component Analysis (PCA), we project the high-dimensional feature spaces into visualizable 2D representations, enabling direct comparison of how different descriptors encode geometric information.
            Our analysis focuses on two key aspects: (1) the clustering behavior of features extracted from patches with similar geometric properties, and (2) the continuity of feature representations during geometric transitions between different surface types. We design two experiments as follows:
            \begin{enumerate}
                \item \textbf{Geometric Type Clustering:} We generate a diverse set of 3D patches that are sampled from four fundamental types of local surface geometry: spherical, parabolic, hyperbolic, and planar patches. To make the clustering task more challenging and better reflect real-world conditions, we add Gaussian noise to the $z$-coordinates of the sampled points.
                \item \textbf{Geometric Transition Analysis:} We create a sequence of 3D patches that smoothly interpolate between two types of local surface geometry, from hyperbolic geometry to spherical geometry.
            \end{enumerate}
            
            For both experiments, we generate 3D patches by sampling polynomials as described above, and also ensuring that the origin $(0,0)$ is included in each patch to serve as a reference point for feature evaluation across all axiomatic methods (WKS, HKS, and SHOT).

            To compute our proposed neural descriptors for each patch, we process its point-cloud through our pipeline to obtain a 2048-dimensional feature vector per patch. For the spectral descriptors (HKS and WKS), we first construct a mesh for each patch, by applying Delaunay triangulation to the patch points projected onto the parametric domain $(x,y)$, then compute\footnote{\href{https://github.com/RobinMagnet/pyFM}{WKS/HKS Implementation}} its feature vector at the origin point $(0,0,f(0,0))$. Similarly, for SHOT features, we utilize the Point Cloud Library's implementation\footnote{\href{https://github.com/humanpose1/python_shot}{SHOT Implementation}} to compute the descriptor at the origin, taking advantage of the local point cloud structure (e.g. for conducting normals estimation, which is essential for SHOT).

\begin{figure}[t]
                \centering
                \includegraphics[width=1.1\textwidth]{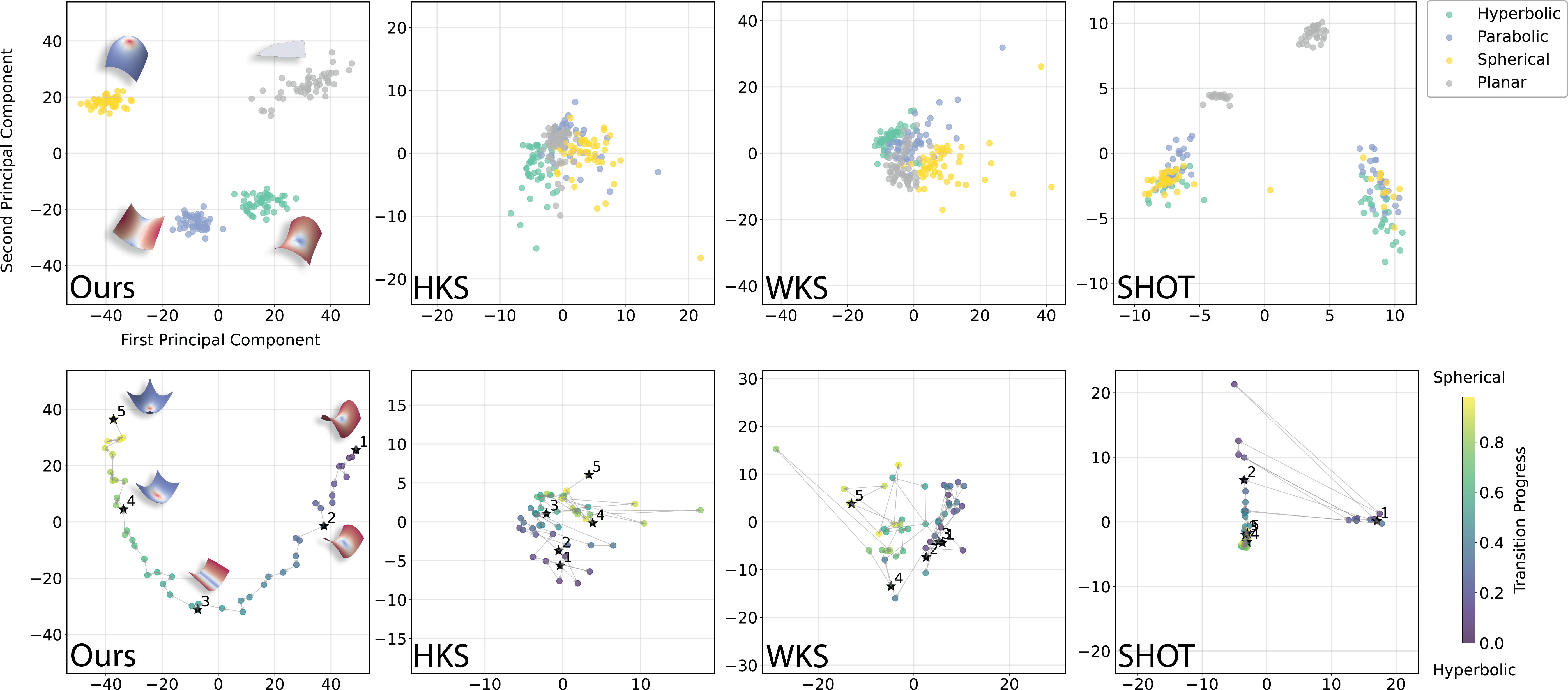}
                \caption{\textbf{PCA visualization of feature spaces.} \textbf{Top Row:} Our features maintain distinct clusters despite noise, while classical descriptors show dispersed, overlapping clusters. \textbf{Bottom Row:} Our features capture smooth geometric transitions with a well-ordered path, while traditional descriptors (HKS, WKS, and SHOT) produce discontinuous trajectories.
                Our visualization approach is inspired by the feature space representation techniques demonstrated in \cite{rotstein2024pathways}.}
                \label{fig:features_pca}
            \end{figure}

            The results of both experiments are visualized in Figure \ref{fig:features_pca}. The top row demonstrates how different feature descriptors maintain cluster coherence in the presence of noise, revealing their stability under realistic scanning conditions. The bottom row shows the feature space trajectories during geometric transitions, illustrating how each descriptor captures the continuous transformation between surface types. Our neural descriptors exhibit both tighter clustering and smoother transitions compared to traditional descriptors.

    

        \subsection{Neural Descriptors and Axiomatic Shape Correspondence}
            We evaluate our method's robustness using the SHREC'16 dataset, specifically focusing on the challenging scenario of establishing correspondences between complete shapes and their versions with holes under isometric deformations. This setup simultaneously tests two critical aspects of our feature descriptors: their invariance to isometric deformations and stability under missing surface regions.
            
            For evaluation, we randomly selected 20 shape pairs and computed both our neural descriptors and traditional descriptors at each vertex. Initial correspondences were established through nearest-neighbor matching in feature space, followed by refinement using the ZoomOut \cite{ZoomOut2019} algorithm. This two-stage approach allows us to assess both the raw discriminative power of the features and their effectiveness as initialization for spectral refinement methods.
            
\begin{figure}[t]
                \centering
                \includegraphics[width=0.75\textwidth]{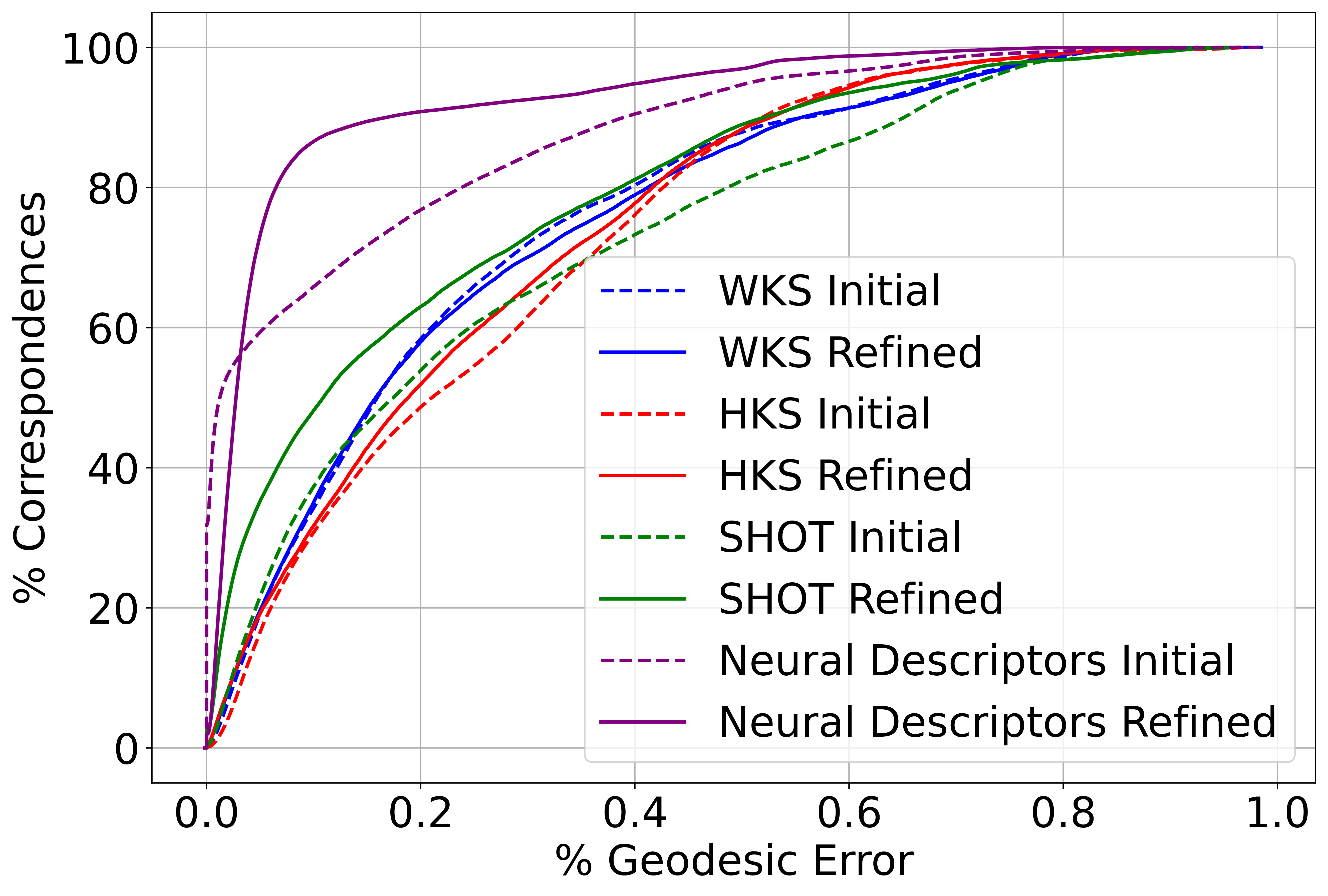}
                \caption{\textbf{Correspondence accuracy comparison on SHREC'16 partial shapes.} Our method (purple) achieves better geodesic error rates compared to traditional descriptors.}
                \label{fig:geodesic_error}
            \end{figure}

            As shown in Figure \ref{fig:geodesic_error}, our neural descriptors significantly outperform traditional descriptors both before and after ZoomOut refinement, demonstrating better preservation of intrinsic geometric properties even in the presence of significant topological noise.
            The plots show the percentage of correspondences with geodesic error below varying thresholds, where higher curves indicate better performance.
            The superior performance of our features is particularly notable in the low-error regime, indicating their ability to establish more precise correspondences.

        \subsection{Neural Descriptors and Learning-Based Shape Correspondence}
            In this section, we integrate our learned neural descriptors into RobustFMNet, a state-of-the-art shape correspondence architecture proposed by Cao et al. \cite{Cao2023}, and repeat their experiments on the FAUST and SCAPE datasets. RobustFMNet jointly optimizes for functional maps and point-wise maps in an unsupervised manner, leveraging a coupling loss to relate the two representations.
            
            To integrate our features with RobustFMNet, we first need to compute neural descriptors for a given 3D mesh. For each vertex on the mesh, we extract a local 3D patch by considering its K-Nearest Neighbors ($K=64$). These patches are then processed through our pipeline to obtain per-vertex neural descriptors. Finally, we feed both the mesh and its computed neural descriptors into RobustFMNet.
            
            By replacing the traditional WKS descriptors used in the original RobustFMNet pipeline with our neural descriptors, we can directly assess the performance and effectiveness of our learned features within this state-of-the-art shape matching system.

            \subsubsection{Evaluation on Isometric Shape Collections.}
                As shown in Table \ref{tab:evaluation_on_SCAPE_and_FAUST}, training RobustFMNet on the FAUST and SCAPE datasets using our neural descriptors as input features achieves state-of-the-art performance. Specifically, we report mean geodesic errors of 1.4, 1.8, 1.5, and 1.7 for the FAUST-FAUST, FAUST-SCAPE, SCAPE-FAUST, and SCAPE-SCAPE test cases, respectively. These results demonstrate a significant improvement over the original RobustFMNet framework \cite{Cao2023} using WKS descriptors, with a relative improvement of up to 12.5\% on FAUST-FAUST and 10.5\% on SCAPE-SCAPE. This highlights the superior ability of our learned descriptors to capture the intrinsic geometric properties of the shapes, enabling more accurate correspondences even in the presence of complex non-rigid deformations.

    
    
    
                \begin{table}[t]
                    \caption{Evaluation on RobustFMNet pipeline on FAUST and SCAPE datasets. For the complete table see \cite{Cao2023}.}
                    \label{tab:evaluation_on_SCAPE_and_FAUST}
                    \centering
                    \begin{tabular}{l|cccc|cccc}
                    \hline
                    \multirow{1}{*}{Train} & \multicolumn{4}{c|}{FAUST} & \multicolumn{4}{c}{SCAPE} \\
                    Test & F & F\_a & S & S\_a & F & F\_a & S & S\_a \\
                    \hline
                    RobustFMNet - WKS - 25 refinement & 1.6 & \textbf{1.9} & 2.2 & \textbf{2.1} & 1.6 & \textbf{2.0} & 1.9 & 2.0 \\
                    RobustFMNet - Neural Descriptors & \textbf{1.4} & 9.5 & \textbf{1.8} & 10.5 & \textbf{1.5} & 10.3  & \textbf{1.7} & 3.3 \\
                    \hline
                    \end{tabular}
                \end{table}


    
    \subsubsection{Robustness to Topological Noise.}
        We evaluate our method on the TOPKIDS dataset to assess robustness to topological noise. As shown in Table \ref{tab:topological_noise_table}, by integrating our neural descriptors into the RobustFMNet framework, we achieve a mean geodesic error of 4.4, marking a substantial 52.2\% improvement over their WKS-based baseline (9.2). This improvement demonstrates our features' superior ability to maintain geometric consistency even in the presence of severe topological perturbations.

        \begin{table}[t]
            \caption{Robustness to topological noise demonstrated on the TOPKIDS dataset. 
            For a complete table with comparisons to additional supervised and unsupervised methods, see \cite{Cao2023}.
            }
            \label{tab:topological_noise_table}
            
            \centering
            \begin{tabular}{lccc}
            \hline
            Geo. error ($\times$100) & TOPKIDS & \\
            \hline
            RobustFMNet - WKS & 9.2  \\
            RobustFMNet - Neural Descriptors & \textbf{4.4} \\
            \hline
            \end{tabular}
        \end{table}

            
    
        \subsubsection{Robustness to Holes and Partial Correspondence.}
            For evaluation on partial shapes, we reevaluate the experiments conducted in \cite{Cao2023} on the SHREC'16 Partial Correspondence benchmark. This dataset includes two challenging scenarios: CUTS (shapes with sharp boundaries from partial shape removal) and HOLES (shapes with multiple missing regions). Our evaluation protocol tests both within-category matching (e.g., training on CUTS and testing on CUTS, referred to as CUTS-CUTS, and similarly for HOLES-HOLES) and cross-category generalization (e.g., training on CUTS and testing on HOLES, referred to as CUTS-HOLES).
            
            As shown in Table \ref{tab:shrec}, our neural descriptors achieve state-of-the-art performance across most scenarios. On the CUTS dataset, we achieve a mean geodesic error of 1.2 after refinement, substantially outperforming other methods. Notably, we achieve these results without any pretraining, in contrast to \cite{Cao2023}, where pretraining was crucial for performance.

        \subsubsection{Qualitative Analysis through Texture Transfer Visualization.}
\begin{figure}[t]
            \centering
            \includegraphics[width=\textwidth]{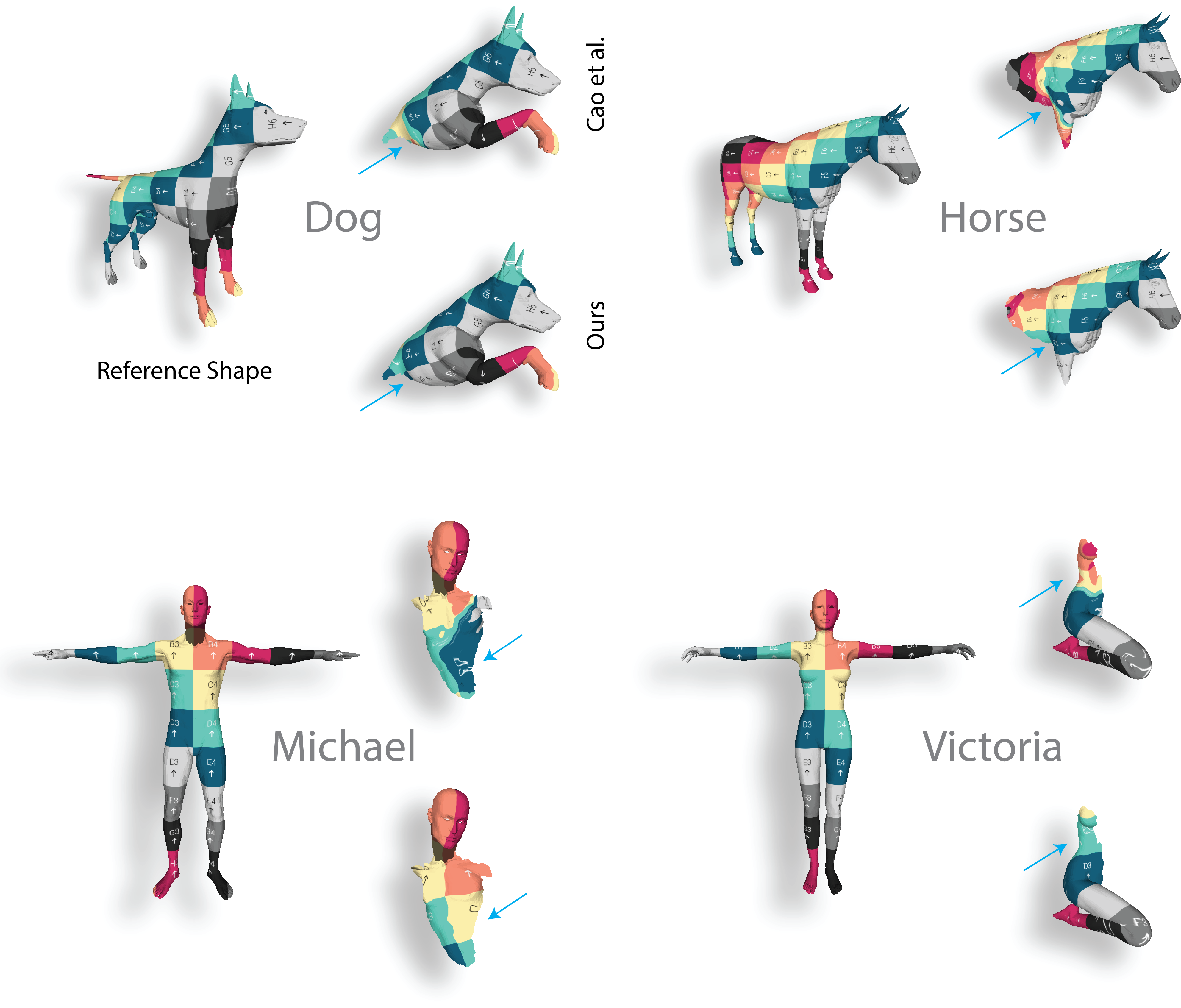}
            \caption{Texture transfer on partial shapes. Follow the blue arrows for visual comparison. Zoom in for details.}
            \label{fig:shrec16_texture_trans}
        \end{figure}

\begin{figure}[t]
            \centering
            \includegraphics[width=\textwidth]{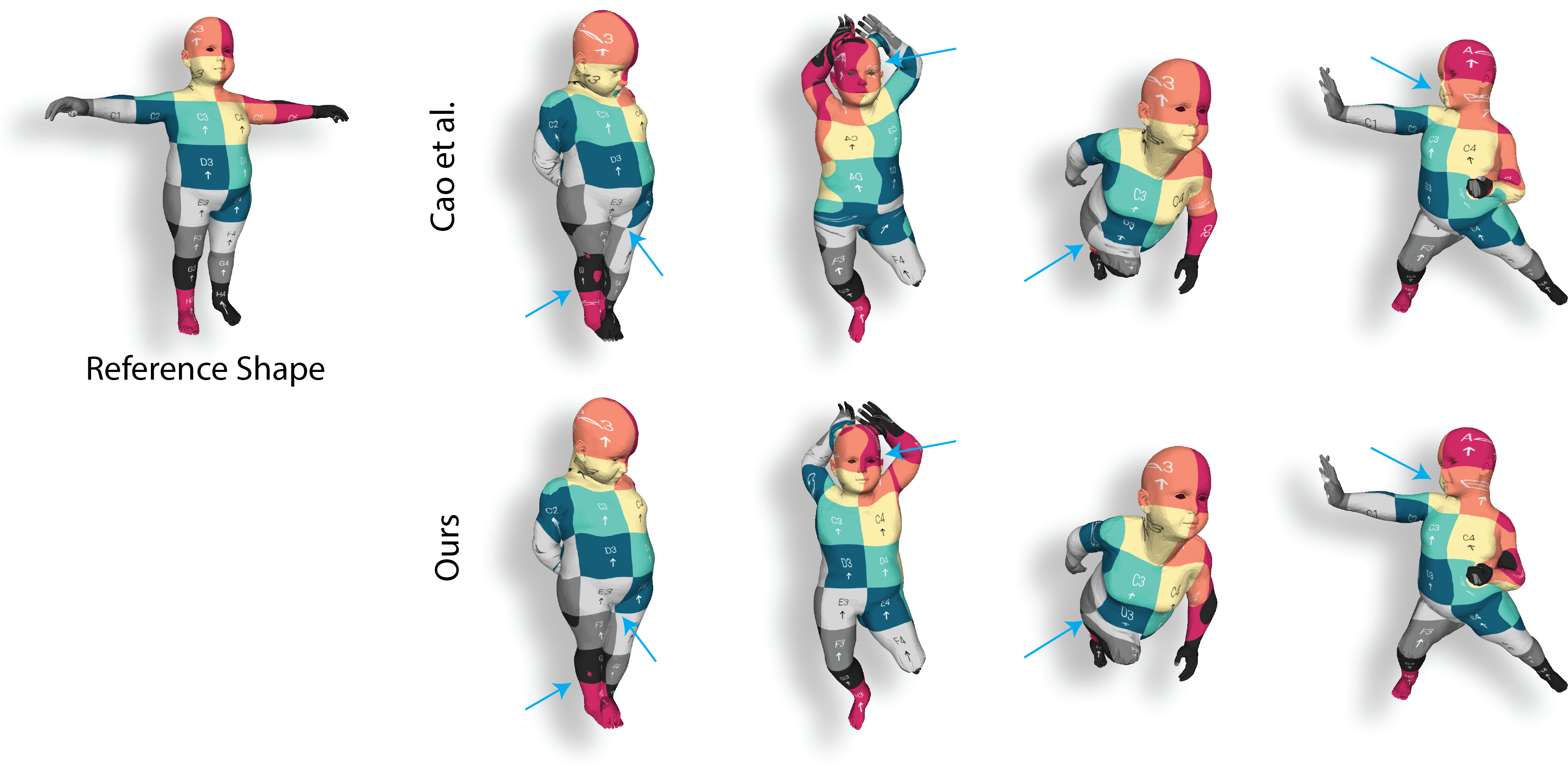}
            \caption{Texture transfer in presence of topological noise. Follow the blue arrows for visual comparison. Zoom in for details.}
            \label{fig:_texture_trans}
            
        \end{figure}

            We visualize the performance of our method through texture transfer, and compare with the results published by Cao et al. \cite{Cao2023}. By mapping a source texture to the target shape, we can assess the accuracy and smoothness of the matching.
            On partial shapes (SHREC'16), our neural descriptors maintain more coherent textures than the WKS baseline used by Cao et al., especially near holes (Fig. \ref{fig:shrec16_texture_trans}).
            Similarly, on shapes with topological noise (TOPKIDS), our approach produces smoother and more accurate texture transfers compared to Cao et al.'s WKS-based method (Fig. \ref{fig:_texture_trans}).
            These qualitative results complement our quantitative evaluations, providing visual confirmation of our method's advantages over the state of the art.

        \begin{table}[t]
            \centering
            \caption{Quantitative evaluation on SHREC'16 partial shape matching benchmark. For the complete table see \cite{bracha2024wormholelosspartialshape}.}
            \label{tab:shrec}
            \begin{tabular}{l|cc|cc}
            Train & \multicolumn{2}{c|}{CUTS} & \multicolumn{2}{c}{HOLES} \\
            Test & CUTS & HOLES & CUTS & HOLES \\
            \hline
            RobustFMnet \cite{Cao2023} - XYZ $\rightarrow$ Refined & 16.9 $\rightarrow$10.6 & 22.7$\rightarrow$16.6 & 18.7 $\rightarrow$ 16.2 & 23.5 $\rightarrow$ 18.8 \\
            WormholeNet \cite{bracha2024wormholelosspartialshape} Orthogonal $\rightarrow$ Refined & \underline{6.9} $\rightarrow$ 5.6 & 12.2 $\rightarrow$ 8.0 & 
            \underline{14.2} $\rightarrow$ \underline{10.2} & \textbf{11.4} $\rightarrow$ \textbf{7.9} \\
            WormholeNet \cite{bracha2024wormholelosspartialshape} LPF $\rightarrow$ Refined & 7.1 $\rightarrow$ \underline{4.7} & \textbf{8.6} $\rightarrow$ \textbf{5.5} & 16.4 $\rightarrow$ 11.6 & \underline{12.3} $\rightarrow$ \underline{8.6} \\
            RobustFMnet \cite{Cao2023} - Ours  $\rightarrow$ Refined & \textbf{1.8} $\rightarrow$ \textbf{1.2} & 11.6 $\rightarrow$ \underline{6.1} & \textbf{8.3} $\rightarrow$ \textbf{3.3} & 13.6 $\rightarrow$ 8.7\\
            \end{tabular}
        \end{table}

    \section{Conclusions and Limitations}
        We have developed a learning-based approach for extracting robust geometric features from 3D surfaces. By combining synthetic data and self-supervision, our method learns sampling-invariant representations that effectively capture intrinsic shape properties. Our experimental results demonstrate the superiority of our learned features across challenging benchmarks, including shape correspondence, robustness to topological noise, and handling of partial shapes.
        
        Despite these advances, our method's performance degrades on meshes with anisotropic triangulation, as can be noticed in Table \ref{tab:evaluation_on_SCAPE_and_FAUST} (SCAPE\_a and FAUST\_a refer to anisotropically remeshed versions of SCAPE and FAUST). This stems from using k-nearest neighbors for patch extraction, which can cause the same local geometric area to appear differently under varying triangulations. This limitation could potentially be addressed in future work by using geodesic patches or adaptive KNN for local patch extraction.
        
        Nonetheless, this work establishes a new state-of-the-art in geometric feature extraction, with potential to benefit a wide range of 3D shape understanding applications.
    
    

    \bibliographystyle{splncs04}
    \bibliography{mybibliography}

\begin{thebibliography}{10}
\providecommand{\url}[1]{\texttt{#1}}
\providecommand{\urlprefix}{URL }
\providecommand{\doi}[1]{https://doi.org/#1}

\bibitem{WKS2011}
Aubry, M., Schlickewei, U., Cremers, D.: {The wave kernel signature: A quantum mechanical approach to shape analysis}. 2011 IEEE International Conference on Computer Vision Workshops (ICCV Workshops)  \textbf{1},  1626--1633 (11 2011)

\bibitem{BaranDeformationTrans2009}
Baran, I., Vlasic, D., Grinspun, E., Popovi\'{c}, J.: Semantic deformation transfer. ACM Trans. Graph.  \textbf{28}(3) (Jul 2009)

\bibitem{BelongieShapeMatchingObjReco2002}
Belongie, S., Malik, J., Puzicha, J.: Shape matching and object recognition using shape contexts. IEEE Transactions on Pattern Analysis and Machine Intelligence  \textbf{24}(4),  509--522 (2002)

\bibitem{bracha2024unsupervisedpartialshapecorrespondence}
Bracha, A., Dagès, T., Kimmel, R.: On unsupervised partial shape correspondence (2024)

\bibitem{bracha2024wormholelosspartialshape}
Bracha, A., Dagès, T., Kimmel, R.: Wormhole loss for partial shape matching (2024)

\bibitem{SIHKS2010}
Bronstein, M.M., Kokkinos, I.: {Scale-invariant heat kernel signatures for non-rigid shape recognition}. 2010 IEEE Computer Society Conference on Computer Vision and Pattern Recognition pp. 1704--1711 (6 2010)

\bibitem{Cao2022}
Cao, D., Bernard, F.: {Computer Vision – ECCV 2022}. Lecture Notes in Computer Science pp. 55--71 (1 2022)

\bibitem{Cao2023}
Cao, D., Roetzer, P., Bernard, F.: Unsupervised learning of robust spectral shape matching. ACM Transactions on Graphics  \textbf{42}(4), ~15 (August 2023)

\bibitem{cartan1935}
Cartan, E.: La m{\'e}thode du rep{\`e}re mobile, la th{\'e}orie des groupes continus et les espaces g{\'e}n{\'e}ralis{\'e}s. The Mathematical Gazette  (1935)

\bibitem{SimSiam2021}
Chen, X., He, K.: {Exploring Simple Siamese Representation Learning}. 2021 IEEE/CVF Conference on Computer Vision and Pattern Recognition (CVPR)  \textbf{00},  15745--15753 (6 2021)

\bibitem{DinhTextureTrans2005}
Dinh, H.Q., Yezzi, A., Turk, G.: Texture transfer during shape transformation. ACM Trans. Graph.  \textbf{24}(2),  289–310 (Apr 2005)

\bibitem{DeepGeometricFunctionalMaps2020}
Donati, N., Sharma, A., Ovsjanikov, M.: {Deep Geometric Functional Maps: Robust Feature Learning for Shape Correspondence}. 2020 IEEE/CVF Conference on Computer Vision and Pattern Recognition (CVPR)  \textbf{00},  8589--8598 (6 2020)

\bibitem{torch_geometric}
Fey, M., Lenssen, J.E.: Fast graph representation learning with {PyTorch Geometric}. In: ICLR Workshop on Representation Learning on Graphs and Manifolds (2019)

\bibitem{UnsupervisedLearningDenseShapeCorr2019}
Halimi, O., Litany, O., Rodolà, E., Bronstein, A., Kimmel, R.: {Unsupervised Learning of Dense Shape Correspondence}. 2019 IEEE/CVF Conference on Computer Vision and Pattern Recognition (CVPR)  \textbf{00},  4365--4374 (6 2019)

\bibitem{DeepFunctionalMaps2017}
Litany, O., Remez, T., Rodolà, E., Bronstein, A., Bronstein, M.: {Deep Functional Maps: Structured Prediction for Dense Shape Correspondence}. 2017 IEEE International Conference on Computer Vision (ICCV) pp. 5660--5668 (10 2017)

\bibitem{ZoomOut2019}
Melzi, S., Ren, J., Rodolà, E., Sharma, A., Wonka, P., Ovsjanikov, M.: {ZoomOut}. ACM Transactions on Graphics (TOG)  \textbf{38}(6),  1--14 (11 2019)

\bibitem{olver_1995}
Olver, P.J.: Equivalence, Invariants and Symmetry. Cambridge University Press (1995)

\bibitem{OLVER2009}
Olver, P.J.: Differential invariants of surfaces. Differential Geometry and its Applications  \textbf{27}(2),  230--239 (2009)

\bibitem{Olver2018NormalForms}
Olver, P.J.: Normal forms for submanifolds under group actions. In: Kac, V.G., Olver, P.J., Winternitz, P., {\"O}zer, T. (eds.) Symmetries, Differential Equations and Applications. pp. 1--25. Springer International Publishing, Cham (2018)

\bibitem{FunctionalMaps2012}
Ovsjanikov, M., Ben-Chen, M., Solomon, J., Butscher, A., Guibas, L.: {Functional maps: a flexible representation of maps between shapes}. ACM Transactions on Graphics  \textbf{31}(4),  1--11 (7 2012)

\bibitem{PishchulinShapeScape2017}
Pishchulin, L., Wuhrer, S., Helten, T., Theobalt, C., Schiele, B.: Building statistical shape spaces for 3d human modeling. Pattern Recogn.  \textbf{67}(C),  276–286 (Jul 2017)

\bibitem{5995486}
Raviv, D., Bronstein, M.M., Bronstein, A.M., Kimmel, R., Sochen, N.: Affine-invariant diffusion geometry for the analysis of deformable 3d shapes. In: CVPR 2011. pp. 2361--2367 (June 2011)

\bibitem{rotstein2024pathways}
Rotstein, N., Yona, G., Silver, D., Velich, R., Bensaïd, D., Kimmel, R.: Pathways on the image manifold: Image editing via video generation (2024)

\bibitem{Roufosse2019}
Roufosse, J.M., Sharma, A., Ovsjanikov, M.: {Unsupervised Deep Learning for Structured Shape Matching}. 2019 IEEE/CVF International Conference on Computer Vision (ICCV)  \textbf{00},  1617--1627 (11 2019)

\bibitem{SHOT2014}
Salti, S., Tombari, F., Stefano, L.D.: {SHOT: Unique signatures of histograms for surface and texture description}. Computer Vision and Image Understanding  \textbf{125},  251--264 (8 2014)

\bibitem{DiffusionNet2022}
Sharp, N., Attaiki, S., Crane, K., Ovsjanikov, M.: {DiffusionNet: Discretization Agnostic Learning on Surfaces}. ACM Transactions on Graphics  \textbf{41}(3),  1--16 (3 2022)

\bibitem{HKS2009}
Sun, J., Ovsjanikov, M., Guibas, L.: {A Concise and Provably Informative Multi‐Scale Signature Based on Heat Diffusion}. Computer Graphics Forum  \textbf{28}(5),  1383--1392 (7 2009)

\bibitem{VelichKimmel2022}
Velich, R., Kimmel, R.: Deep signatures - learning invariants of planar curves (2022)

\bibitem{VelichKimmel2023}
Velich, R., Kimmel, R.: Learning differential invariants of planar curves. In: Calatroni, L., Donatelli, M., Morigi, S., Prato, M., Santacesaria, M. (eds.) Scale Space and Variational Methods in Computer Vision. pp. 575--587. Springer International Publishing, Cham (2023)

\bibitem{DeltaConv2022}
Wiersma, R., Nasikun, A., Eisemann, E., Hildebrandt, K.: {DeltaConv}. ACM Transactions on Graphics  \textbf{41}(4),  1--10 (7 2022)

\bibitem{PointTransformer2021}
Zhao, H., Jiang, L., Jia, J., Torr, P., Koltun, V.: {Point Transformer}. 2021 IEEE/CVF International Conference on Computer Vision (ICCV)  \textbf{00},  16239--16248 (10 2021)

\end{thebibliography}
\end{document}